\documentclass[conference]{IEEEtran}
\IEEEoverridecommandlockouts
\usepackage{stfloats}
\usepackage{cite}
\usepackage{amsmath,amssymb,amsfonts}
\usepackage{algorithmic}
\usepackage{graphicx}
\usepackage{textcomp}
\usepackage{xcolor}
\usepackage{tabularx}
\usepackage{csquotes}
\usepackage{graphicx}
\usepackage{subcaption}
\usepackage{url}
\usepackage{float}   
\usepackage{booktabs} 

\usepackage{fancyhdr}
\fancypagestyle{withfooter}{

\fancyfoot[C]{\footnotesize Accepted to the Novel Approaches for Precision Agriculture and Forestry with Autonomous Robots IEEE ICRA Workshop - 2025}
}
\AtBeginEnvironment{quote}{\itshape}
\def\BibTeX{{\rm B\kern-.05em{\sc i\kern-.025em b}\kern-.08em
    T\kern-.1667em\lower.7ex\hbox{E}\kern-.125emX}}
\begin{document}
\newcommand{\ebasa}[1]{\textit{\textcolor{red}{{Ebasa: }#1}}}
\newcommand{\sarah}[1]{\textit{\textcolor{blue}{{Sarah: }#1}}}
\newcommand{\mario}[1]{\textit{\textcolor{orange}{{Mario: }#1}}}
\title{Geofenced Unmanned Aerial Robotic Defender for Deer Detection and Deterrence (GUARD)\\

}

\author{
    \IEEEauthorblockN{%
        Ebasa Temesgen\IEEEauthorrefmark{1},
        Mario Jerez\IEEEauthorrefmark{1},
        Greta Brown\IEEEauthorrefmark{1},
        Graham Wilson\IEEEauthorrefmark{2},\\
        Sree Ganesh Lalitaditya Divakarla\IEEEauthorrefmark{1},
        Sarah Boelter\IEEEauthorrefmark{1},
        Oscar Nelson\IEEEauthorrefmark{2},\\
        Robert McPherson\IEEEauthorrefmark{2},
        Maria Gini\IEEEauthorrefmark{1}}
    \IEEEauthorblockA{\IEEEauthorrefmark{1}Dept.\ of Computer Science \& Engineering, University of Minnesota\\
                     \IEEEauthorrefmark{2}Dept.\ of Mechanical Engineering, University of Minnesota\\
                     Minneapolis, MN, USA\\
                     \{temes021, jerez005, brow6802, divak014, boelt072, gini\}@umn.edu, \{wils3101, nel13668, mcphe258\}@umn.edu}
}

\maketitle

\begin{abstract}
Wildlife-induced crop damage, particularly from deer, threatens agricultural productivity. Traditional deterrence methods often fall short in scalability, responsiveness, and adaptability to diverse farmland environments. This paper presents an integrated unmanned aerial vehicle (UAV) system designed for autonomous wildlife deterrence, developed as part of the Farm Robotics Challenge. Our system combines a YOLO-based real-time computer vision module for deer detection, an energy-efficient coverage path planning algorithm for efficient field monitoring, and an autonomous charging station for continuous operation of the UAV. In collaboration with a local Minnesota farmer, the system is tailored to address practical constraints such as terrain, infrastructure limitations, and animal behavior. The solution is evaluated through a combination of simulation and field testing, demonstrating robust detection accuracy, efficient coverage, and extended operational time. The results highlight the feasibility and effectiveness of drone-based wildlife deterrence in precision agriculture, offering a scalable framework for future deployment and extension. 
\end{abstract}


\section{Introduction}

Crop damage caused by wildlife, particularly deer incursions, represents a challenge for modern agriculture. Deer damage to crops is responsible for disagreements among farmers, hunters, and the Department of Natural Resources over how the deer population should be controlled \cite{kennedy_minnesota_2024}. The prevalence of deer leads to substantial reductions in crop yield and increased operational costs. White-tailed-deer depredation now costs Midwestern growers \$8–10 M per year in direct yield loss and mitigation expenses \cite{tyndall_deerstudy_2024}.
Traditional methods of wildlife deterrence, such as fences and scare tactics, often lack the responsiveness and adaptability required to protect expansive and heterogeneous farmlands, but commercial deterrence methods for large scale farms are often very expensive or too large for small farms.

While permanent exclusion fencing is a widely adopted approach to deer deterrence, it is rarely economical for small holdings.  High-tensile woven-wire systems cost about \$10\ per ft. to install—roughly \$26,000 for a 0.4 km perimeter \cite{generutgers}.  

Gates, machinery access and annual maintenance compound these drawbacks, making static fencing impractical for the 13-acre vegetable farm that motivated this study.  Alternative auditory and visual repellents are only transiently effective: field experiments record complete habituation within two to four weeks \cite{valitzski2007evaluation}.  By comparison, low-altitude multirotors have been shown to displace large mammals more reliably than ground-based hazing methods \cite{sarmento2025drones}.  These considerations motivated a mobile, autonomous and cost-aware solution.

This work was initiated as part of our participation in the \emph{Farm Robotics Challenge}\footnote{\url{https://www.farmroboticschallenge.ai/}}, a competition designed to encourage innovative robotic solutions addressing real-world agricultural issues. Our team partnered with a local farmer operating a 13-acre vegetable farm located just east of Woodbury, Minnesota. When discussing the challenges posed by deer, the farmer shared detailed insights:

\blockquote{ We plant things out into the field in the middle of April—at the earliest—with the last harvest around the end of October. Deer start eating crops as soon as they find them, forming eating habits that keep them returning frequently. Nighttime, particularly around dusk, is when deer are most active, though we also see them during the day occasionally, especially in the spring. I've noticed deer regularly visiting our fields, and whenever they're scared off, they don't just linger next door—they run to other areas they remember have food.}

Understanding there were also cost constraints for small farms, we chose to develop an integrated unmanned aerial vehicle (UAV) system designed specifically for autonomous wildlife deterrence. We utilize low-cost hardware for the sytem. The system utilizes advanced computer vision algorithms, specifically the You Only Look Once (YOLO)\cite{yolo} architecture, for robust deer detection under diverse and dynamic field conditions. Additionally, a generic coverage path planning algorithm ensures efficient and extensive monitoring across the entire farmland.

Unmanned aerial vehicles (UAVs) are increasingly accessible for wildlife monitoring, yet robust perception remains difficult.  
Axford et al.~\cite{axford2024collectively} surveyed deep-learning methods for aerial wildlife detection, identifying persistent challenges such as occlusion, small targets, dense vegetation and limited annotated data.  
Crossling et al.~\cite{crossling2024automated} showed that models trained on clean stock photos (VGG16, ResNet50) generalise poorly to on-farm imagery, underscoring the need for domain-matched datasets and simulation.  
To bridge this gap, we developed a Airsim \cite{shah2018airsim} based simulator built on top of Unreal Engine that procedurally generates synthetic farmland scenes with photorealistic deer models, enabling training for a reinforcement-learning (RL) agent that performs deterrence behaviour.  
Integration and field validation of this agent are ongoing; the present paper reports the hardware platform, perception stack and baseline coverage controller on which the RL policy will execute.




We used a PX4 based quad-copter equipped with an NVIDIA Jetson Orin Nano for onboard inference.  
A fine-tuned YOLOv5 detector \cite{yolov5} provides real-time deer localisation, while a learning-aided coverage-path planner ensures energy-efficient monitoring of irregular fields. We also propose a proof of concept for a contact-based charging dock near the farmhouse to extend dusk-to-dawn sorties without human intervention.

The primary contributions of this paper include:
\begin{itemize}
\item A hardware and software integration for autonomous wildlife deterrence tailored to practical agricultural constraints.
\item Implementation of an optimized YOLO-based computer vision module for real-time deer detection and tracking.
\item Development of a versatile coverage path planning algorithm capable of energy-efficient coverage of agricultural fields.
\item Integration of an autonomous charging mechanism, enabling prolonged UAV operations with minimal human intervention.
\end{itemize}

\section{Related Work}

\subsection{Drone Applications Pest Deterrence}
Some animals are may cause damage or annoyance in difficult to reach places during inconvenient times, leaving deterrence by drones as the optimal option. Schiano et al. \cite{f_schiano_autonomous_2022} discuss urban pigeon deterrence, as they frequently transmit diseases to humans and cause damage to infrastructure. Schiano et al. used a video camera with a neural network mounted on a roof and utilizing a drone for pigeon deterrence, the drone was autonomously deployed 55 times successfully to deter pigeons over a 21 day period. In addition to urban spaces, birds also cause significant damage to crops. Wang et al. \cite{z_wang_autonomous_2019} proposed an algorithm for bird deterrence using multiple UAVs and a bird behavior created from field trials. Autonomous cooperative missions were more effective than a single UAV for bird deterrence from crops.

\subsection{Computer Vision for Animal Detection}
Recent advances in UAV technology have spurred significant research in agricultural robotics, particularly in the areas of crop monitoring, pest control, and wildlife deterrence. 
There are various methods for detecting tagged and untagged wildlife and livestock. Cliff et al. \cite{cliff_online_nodate} utilized drones with antennae for localizing radio-tagged wildlife. Since we will primarily be deterring deer, we don't have the constraint or focal point of tag tracking a specific group of animals. Additionally, existing methods utilizing radio frequency or visual ear tags are prone to loss or damage. Using methods with video or photo analysis will likely fit our use case better since we monitor a specific area. Qiao et al. developed a deep learning approach for cattle identification using video analysis \cite{qiao_automated_2021}. Also in the field of livestock management, Brown et al. \cite{brown_automated_2022} explored methodologies to use the lowest resolution drone images possible while still being able to identify livestock accurately to allow farmers and ecologists to use more accessible medium resolution imagery for tasks. 

Tracking deer specifically requires more planning. Deer tend to be active at sunrise and after sunset, and may not easily be tracked in dark or hazy environments. Logan et al. \cite{logan_improving_2019} discuss the utility of thermal imaging to detect deer during night hours to aid culling and management. Surveillance methods can also help ensure animal safety and disease management. Rodriguez \cite{rodriguez_deer_nodate} discusses deer surveillance in public spaces. A YOLOv5 object detection algorithm was used and the model was effectively trained to recognize people and deer, as well as to provide a document with the labels and the coordinates of the object detected in the image for further processing.

\subsection{Autonomous Charging Systems for UAV}
Autonomous charging mechanisms have also been explored to extend UAV operation times in field deployments. An issue with using patrolling UAVs for deterrence of animals is battery life. Even if UAVs patrol only when motion or animals are detected, the battery may only last a few laps of a farm perimeter before having to be manually changed out or recharged. Ure et al. \cite{ure_automated_2015} discuss a compact autonomous battery maintenance system that significantly extends the operational time of battery powered UAVs. The rotational design allowed a persistent 3 hour mission with 100 battery swaps. Nieuwoundt et al. \cite{h_nieuwoudt_automated_2023} utilized battery recharging for drones instead of swapping using a docking station and were able to successfully center the drones on the landing platform for recharging. 

Despite these advances, few works have integrated all these components into a single, cohesive system capable of robust real-world performance. Our work aims to bridge this gap by presenting an integrated approach validated in a competitive setting.

\begin{figure*}[ht]{
  \includegraphics[width=\textwidth]{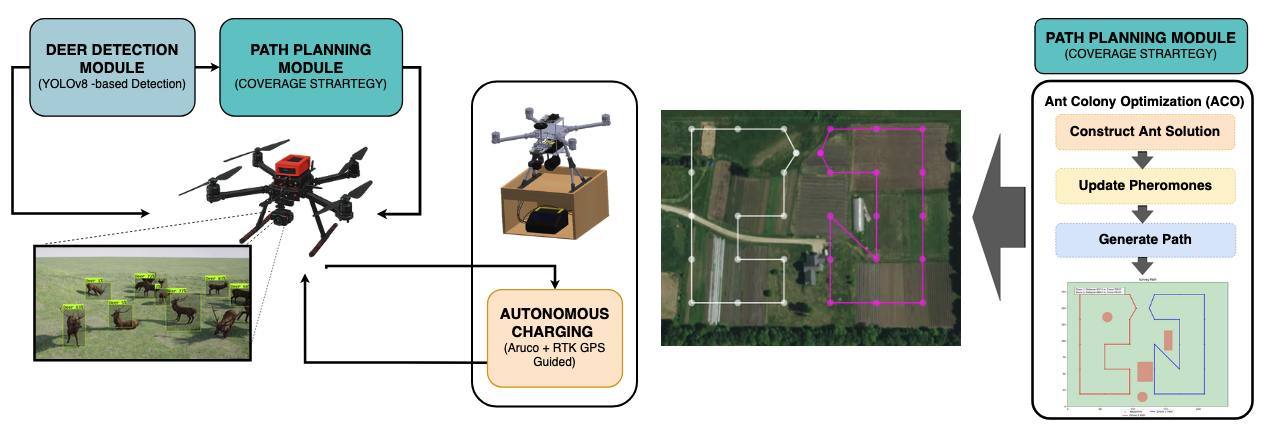}
\caption{ High-level system architecture showing integration of computer vision, path planning, and autonomous charging modules.}
\label{fig:system_architecture}
}
\end{figure*}
\section{System Overview and Architecture}


Our solution is an integrated UAV platform composed of four primary modules for real-time detection, area coverage, charging, and supervision. The UAV itself is a quadcopter outfitted with a camera and onboard computing (e.g., Jetson Orin Nano) for running a YOLO-based vision pipeline. A coverage path planning module determines efficient flight routes to survey the field, while an autonomous charging system handles power maintenance. Overseeing these processes is a supervisory module that coordinates detection triggers, flight planning updates, and battery status checks.

During flight, the quadcopter follows waypoints generated by the path planning module and continuously analyzes the camera feed for deer. Upon detection, the supervisory module can interrupt the default flight plan to direct the UAV toward a deterrence maneuver. Once battery levels fall below a set threshold, the UAV navigates back to the charging station, using an ArUco marker-based docking procedure for precise landing and a balanced recharge cycle. After recharging, the drone automatically returns to its assigned waypoints, resuming field surveillance with minimal downtime.

Figure~\ref{fig:system_architecture} illustrates the high-level system architecture and inter-module communication. This modular design keeps each component self-contained, simplifying debugging, maintenance, and future upgrades—while also ensuring that mission-critical tasks such as detection and deterrence can adapt on the fly to real-world conditions. By unifying advanced computer vision, coverage path planning, and autonomous battery management within a single platform, our system delivers persistent, on-demand deterrence in a robust and scalable manner.

\section{Component Modules}
The quadcopter carries an on board computer running tightly-coupled modules: a YOLO-based vision stack that spots deer in real time, a coverage planner that issues energy-aware waypoints, an autonomous charging routine that handles return-to-dock and. Image frames flow from the camera to the vision module; detections and battery state are passed from the PX4 flight controller to the RL agent, which will decides whether to keep following the coverage route, initiate a deterrence manoeuvre, or head for the dock. 
\subsection{Autonomous Charging}
Our design for the autonomous charging is a proof-of-concept that took several main factors into account: positional repeatability, sufficient electrical contact between the drone and the charging components, and balanced charging of the battery cells. 


 \begin{figure}[ht]
    \centering
    \begin{subfigure}[b]{0.7\linewidth}
        \centering
        \includegraphics[width=\linewidth]{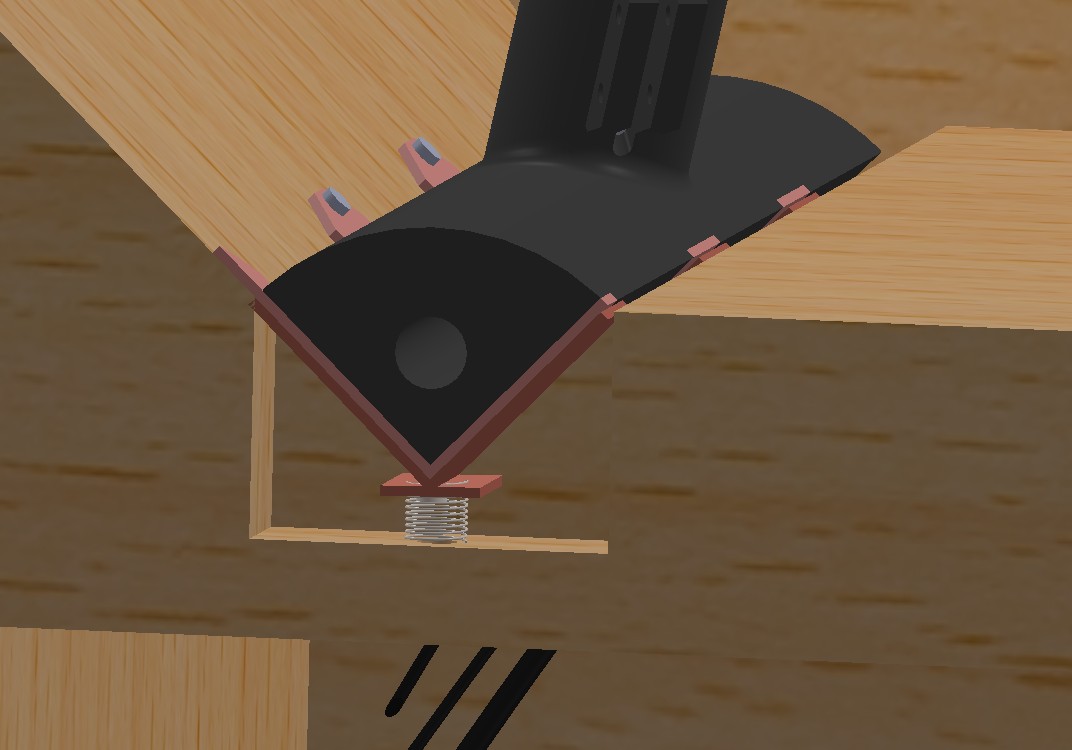}
        \caption{Skid–pad cross-section.}
        \label{fig:landing_c}
    \end{subfigure}\\ 
    \begin{subfigure}[b]{0.7\linewidth}
        \centering
        \includegraphics[width=\linewidth]{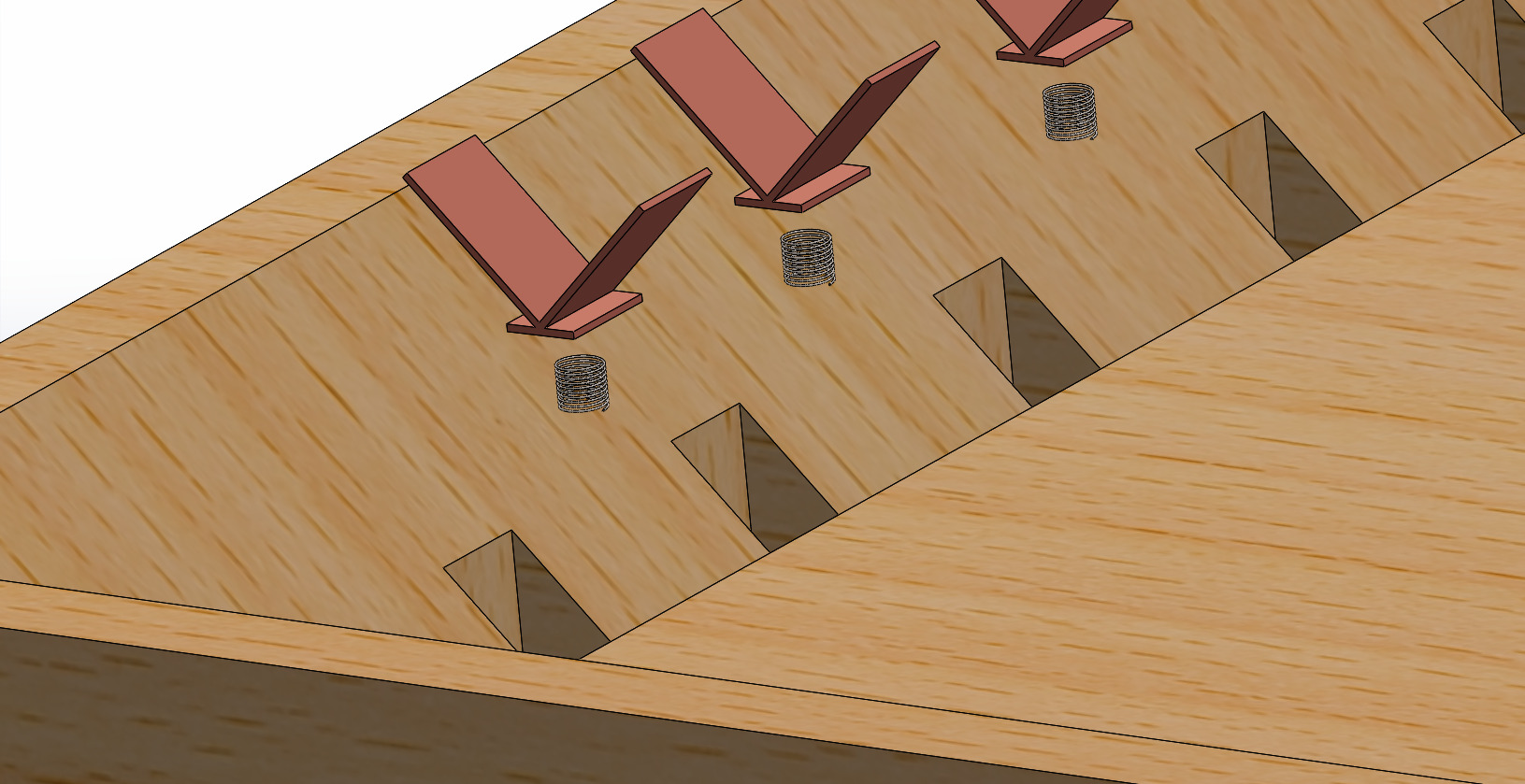}
        \caption{Pad interior with pogo-springs.}
        \label{fig:pad_interior}
    \end{subfigure}
    \caption{Ground-based autonomous charging dock.  
    (a) Inter-locking \(60^{\circ}\) V-rails provide passive self-centering;  
    (b) spring-loaded pogo pins supply power and balance leads;  
    }
    \label{fig:charging_overview}
\end{figure}


The outer profile forms a pair of \(60^{\circ}\) V-rails; complementary V-grooves are milled into a birch-ply pad (Fig.~\ref{fig:landing_c}).  
During descent an RTK-GPS fix brings the UAV to within \(0.5\)m of pad centre; a downward camera then locks on to a 16cm ArUco tag for sub-decimetre alignment, following \cite{moreira2021precision}.  
Any residual \(\leqslant\!3\,\mathrm{cm}\) XY error is passively corrected by the interlocking rails—an approach analogous to the passive funnels reviewed in \cite{galimov2020uav}.  
Empirically the vehicle settles into a unique pose whenever touchdown occurs inside a \(6\)cm radius envelope, matching the tolerance reported for magnetically assisted pads such as AutoCharge \cite{saviolo2023autocharge}.  
The 6-cell pack is charged by an off-the-shelf ISDT Q8 balance charger (0.8 C, 6 A).  
Two power leads and eight balance-tap wires terminate in spring-loaded copper pogo pins embedded in the pad; stainless compression springs provide \(\sim 1.5\) N per pin to maintain \(<50\) m\(\Omega\) contact resistance, an arrangement similar to \cite{kokkinos2023design}.  
Inline 20 A blade fuses protect the rails, while a 75 °C thermal fuse is bonded to the pack.



During the initial design and testing phase of the project we focused on modeling the drone landing gear and pad, as well as working on implementing the autonomous landing using ArUco markers. In the next phase of the project we intend to complete the circuit design for the charging system as well as manufacture a landing pad prototype for testing and design refinement. We also plan to explore other options for ensuring repeatable landings and positively locking the drone landing gear into the pad during charging cycles. CAD models have been validated in Gazebo; plywood prototypes are under construction.





\subsection{Computer Vision Module}

The computer vision system is centered on YOLOv5, a real-time object detection framework optimized for low-latency inference on embedded platforms. Our objective is to detect deer in outdoor environments (as shown in Figure 3) under challenging conditions, including partial occlusions, variable lighting, and natural camouflage, using edge-deployable hardware. The model is deployed on the NVIDIA Jetson Orin Nano, leveraging TensorRT with \texttt{FP16} precision, enabling inference times of less than 25 milliseconds per 640$\times$640 frame.

The model was fine-tuned on a domain-specific annotated dataset of deer, comprising both daylight and infrared night-vision imagery. This dataset was obtained from Roboflow Universe and is tailored to the \textit{Rucervus duvaucelii} (barasingha deer), including scenes representative of wild or semi-structured environments ~\cite{roboflow_dataset}. It includes bounding box annotations for single and multiple deer instances across varied backgrounds, lighting conditions, and poses, ensuring generalizability for real-world deployment.

Post-inference, bounding boxes with confidence scores are generated for each frame. To reduce noise and false detections, a temporal-spatial filtering algorithm is applied. This includes:
\begin{itemize}
    \item IoU-based object tracking for maintaining detection identity across frames
    \item Confidence thresholding, based on F1-optimized values
    \item Geofence-aware rejection, using prior maps of occluded or non-observable areas
\end{itemize}

The deployment scenario includes drone surveillance over agricultural land. Operational maps define UAV boundaries, restricted zones (e.g., tall obstacles), and designated charging stations. This spatial context supports robust filtering of irrelevant detections and false triggers. Fig. \ref{fig:farm} illustrates the birds-eye view of the deployment region.


\subsection{Path Planning Module}

\begin{figure}[ht]
    \centering
    \includegraphics[width=1\linewidth]{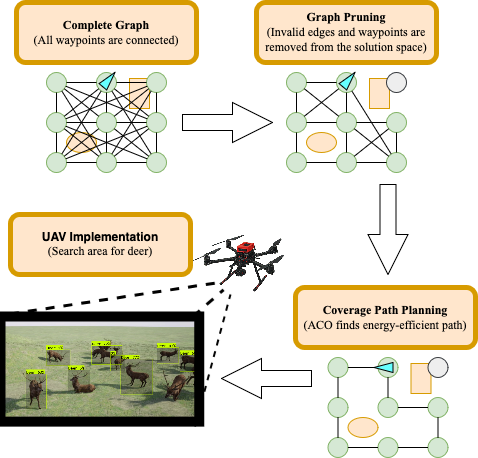}
    \caption{Illustration of our methodology for energy-efficient deer search.}
    \label{fig:aco_arch}
\end{figure}

\begin{figure}[ht]
    \centering
    \includegraphics[width=0.75\linewidth]{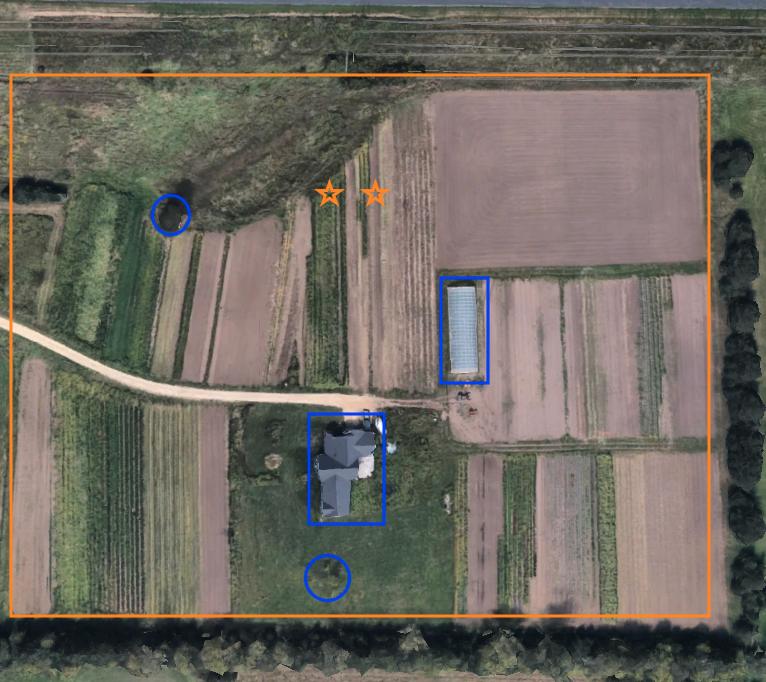}
    \caption{Birds-eye view of the farm our solution is designed for. Within the perimeter (orange rectangle) of the farm, there are obstacles such as tall trees (blue circles), a house and a greenhouse (blue rectangles). Orange stars represent the placement of the UAV charging stations.}
    \label{fig:farm}
\end{figure}

Our path planning module employs a coverage path planning strategy aimed at maximizing area surveillance while minimizing the energy consumption of the UAV. Our approach to generating an efficient coverage path is illustrated by Figure \ref{fig:aco_arch}. We begin by creating a 2D representation of the farm (see figure \ref{fig:farm}), and creating a grid of evenly distributed waypoints throughout the environment. The distance between waypoints is set close enough so that if a drone passes through each waypoint, the drone will have scanned the entire farm through the on-board camera, and far enough apart that there is not excessive overlap in area covered by the drone when traversing neighboring waypoints. In our implementation, we arbitrarily set waypoints 38 meters apart. More work has to be done to determine the ideal spacing between waypoints for our scenario.


We define a solution to the path coverage problem as an ordered array of edges that describes a path that a UAV takes to fly from its charging station, through all its valid waypoints, and back to its charging station. An edge, or a solution component $c_{ij}$ is a straight-line path from node $i$ to node $j$. A valid waypoint is one that is not within some definable distance of any of the collideable structures illustrated in Figure \ref{fig:farm}. The solution space is initially defined as the collection of all possible edges between waypoints. This means that initially, every waypoint is connected to all other waypoints by some edge in the solution space. We then ensure obstacle avoidance by removing edges from the solution space that would cause a UAV to collide with or come close to colliding with an obstacle. This also means that the solution space no longer contains edges connected to invalid waypoints.

The coverage area is divided into sub-areas that can be reasonably covered by a single UAV, and all of the waypoints that are within the bounds of a sub-area are assigned to one drone. Sub-areas should be decided so as to minimize the number of path intersections between different UAVs; however, as it is practical to have landing stations clustered by a power supply, some intersecting paths between drones may be unavoidable, depending on where charging stations are placed. This can be handled by having drones with intersecting flight paths stay at different altitudes \cite{z_wang_autonomous_2019}. In our study, we use this strategy to cover a farm with one drone–the single-drone problem–and with two drones–the dual-drone problem.

We define the heuristic function of each edge $c_{ij}$ to be the inverse of the energy expenditure required to go from waypoint $i$ to waypoint $j$. As research has shown that the distance a drone travels in a linear direction and the amount that it turns are key factors in energy expenditure during a flight path~\cite{j_modares_ub-anc_2017}, we estimate the energy costs of flight paths by these two factors. Thus, we have the estimation of energy cost of a solution $\texttt{cost}(s) = \lambda \cdot d_s + \gamma \cdot \theta_s$, where $\lambda = 0.1164$ is the energy cost of traveling in a straight line in kJ/m and $\gamma = 0.0173$ is the energy cost of turning, given in kJ/deg. We use the values of $\lambda$ and $\gamma$ provided by previous literature that used a drone similar to ours~\cite{j_modares_ub-anc_2017}. $d_s$ and $\theta_s$ are the total distance and rotation that a drone must travel measured in meters and degrees respectively.

\begin{figure}[ht]
    \centering
    \includegraphics[width=0.7\linewidth]{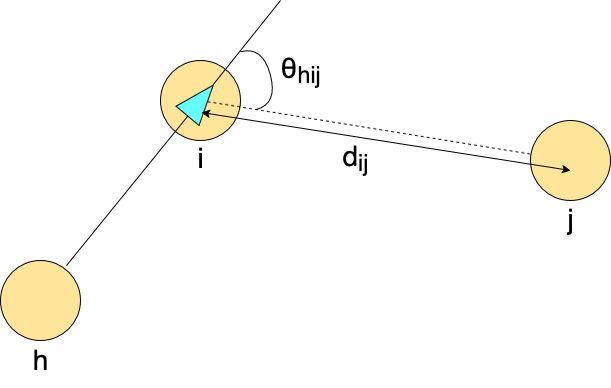}
        \caption{Illustration of the parameters that contribute to energy cost estimation for a drone (blue triangle) to go from waypoint (orange circle) i to waypoint j, given that it was previously on waypoint h. $\theta_{hij}$ is the amount that the drone must turn, and $d_{ij}$ is the linear distance it must travel.}
        \label{fig:cost_variables}
\end{figure}

Finally, we set up a combinatorial optimization problem similar to the Traveling Salesperson Problem (TSP) \cite{lawler1985traveling}, in which a drone begins at its charging station, traverses to every valid waypoint exactly once and returns to its charging station. While in TSP, a path is evaluated according to its total distance, when evaluating a path in our problem formulation, both total distance and rotation are considered. We implement this by setting the heuristic function for a drone to go from waypoint $i$ to waypoint $j$ to be $\eta_{hij} = \frac{1}{\lambda  d_{ij} + \gamma \theta_{hij}}$, such that $h \neq i \neq j$, given that the drone was previously at waypoint h. as illustrated in figure \ref{fig:cost_variables}, $d_{ij}$ is the distance from waypoint i to waypoint j and $\theta_{hij}$ is the positive amount that the drone must rotate to go to waypoint j.

Finally, we use Ant Colony Optimization (ACO)~\cite{colorni_distributed_1991, dorigo_optimization_1992, dorigo_ant_2006} to generate solutions to the problem described. We implement both Ant System (AS)~\cite{dorigo_ant_2006} and Max-Min Ant System (MMAS)~\cite{stutzle_max-min_1997, stutzle_maxmin_2000} ACO which is known to perform strongly when solving the TSP~\cite{v_murugananthan_traveling_2023}. We use a back-and-forth algorithm~\cite{y_cao_concentrated_2022} as a baseline to compare the ACO-generated paths to, as it is commonly used for similar coverage path planning scenarios~\cite{israel_uav-based_2012, k_bezas_coverage_2021, vazquez-carmona_coverage_2022, mu_coverage_2023, r_i_mukhamediev_coverage_2023, torres_coverage_2016}.



\subsection{Deer Deterrence and Control}


Reinforcement learning (RL) is the core of our integration and control strategy, serving as a centralized controller that unifies the vision, path planning, and charging subsystems. Rather than using a static algorithm to orchestrate these modules, we employ a learning-based approach in which the drone becomes an autonomous agent that continually refines its behavior based on interactions with both simulated and real farmland environments. In this paradigm, each drone receives sensory data, such as detection outputs and battery status, and dynamically decides its next action (e.g., reroute, activate deterrence, or return to charge) through a policy derived from RL training.

The RL agent is trained primarily in a photorealistic Airsim built on top of Unreal Engine simulation, where virtual deer roam procedurally generated farmland scenes. This environment provides diverse and repeatable scenarios for the drone to encounter deer under varying lighting, weather, and terrain conditions. The drone’s reward function balances several objectives: maintaining adequate coverage, effectively deterring deer, and avoiding premature battery depletion. The simulation allows robust policy learning that adapts to varying real-world challenges by exposing the agent to thousands of randomized episodes.

Once the simulated policy demonstrates stable performance in accurate deer avoidance/deterrence and efficient battery usage, we will begin incremental real-world testing. We will start with controlled field trials on a small, delineated test plot, using domain randomization strategies (e.g., noise injection for sensor data) to mitigate sim-to-real discrepancies. The drone will collect new data in real-time, refining its policy through additional offline training “updates.” Over multiple iterations, the system will converge toward a policy that responds appropriately to unexpected events, such as sudden appearances of deer or rapidly changing environmental factors, while maintaining safe flight operations.

\begin{figure}
    \centering
    \includegraphics[width=1\linewidth]{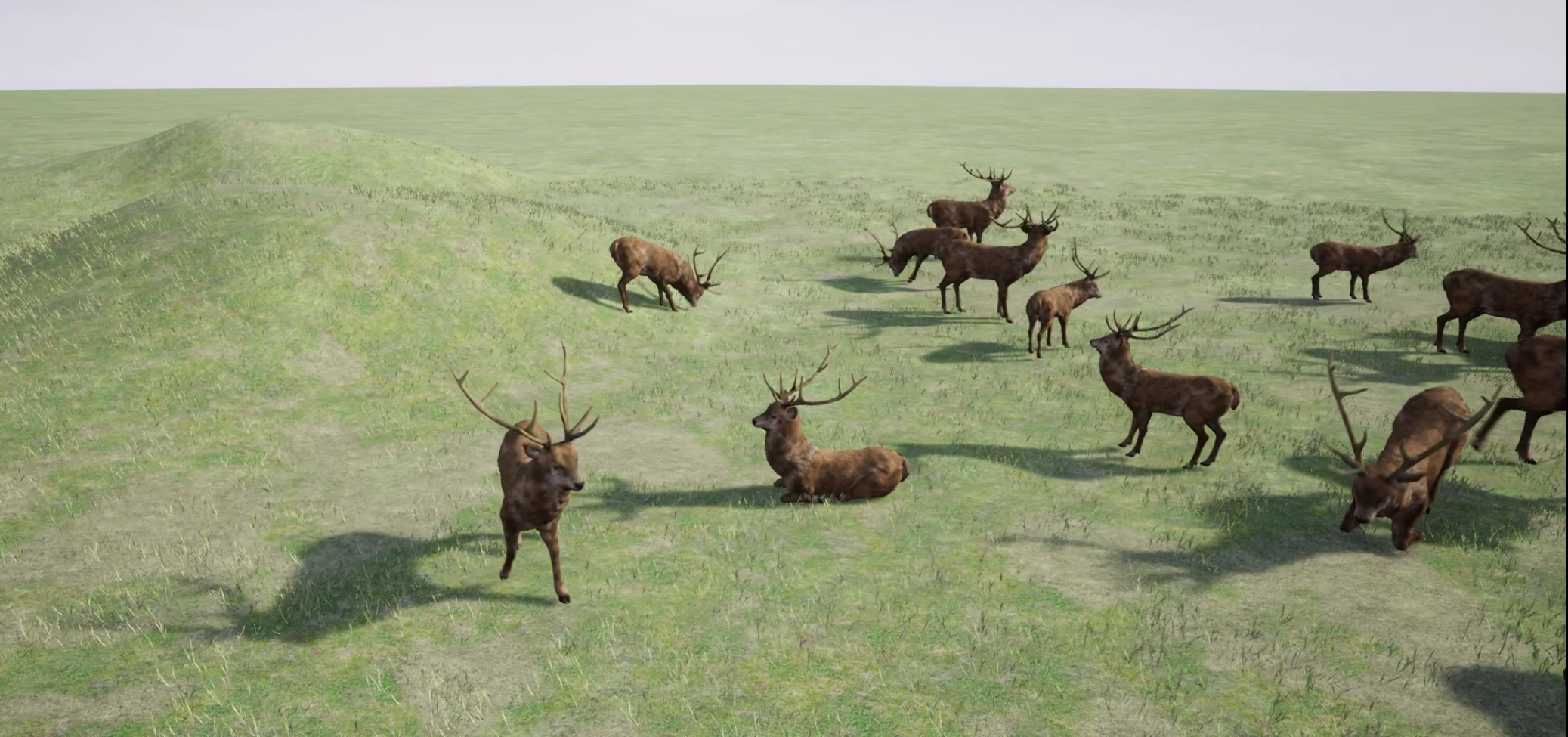}
    \caption{Virtual deer herd with realistic behaviours.}
    \label{fig:deer_behavior}
\end{figure}

\begin{figure}
    \centering
    \includegraphics[width=1\linewidth]{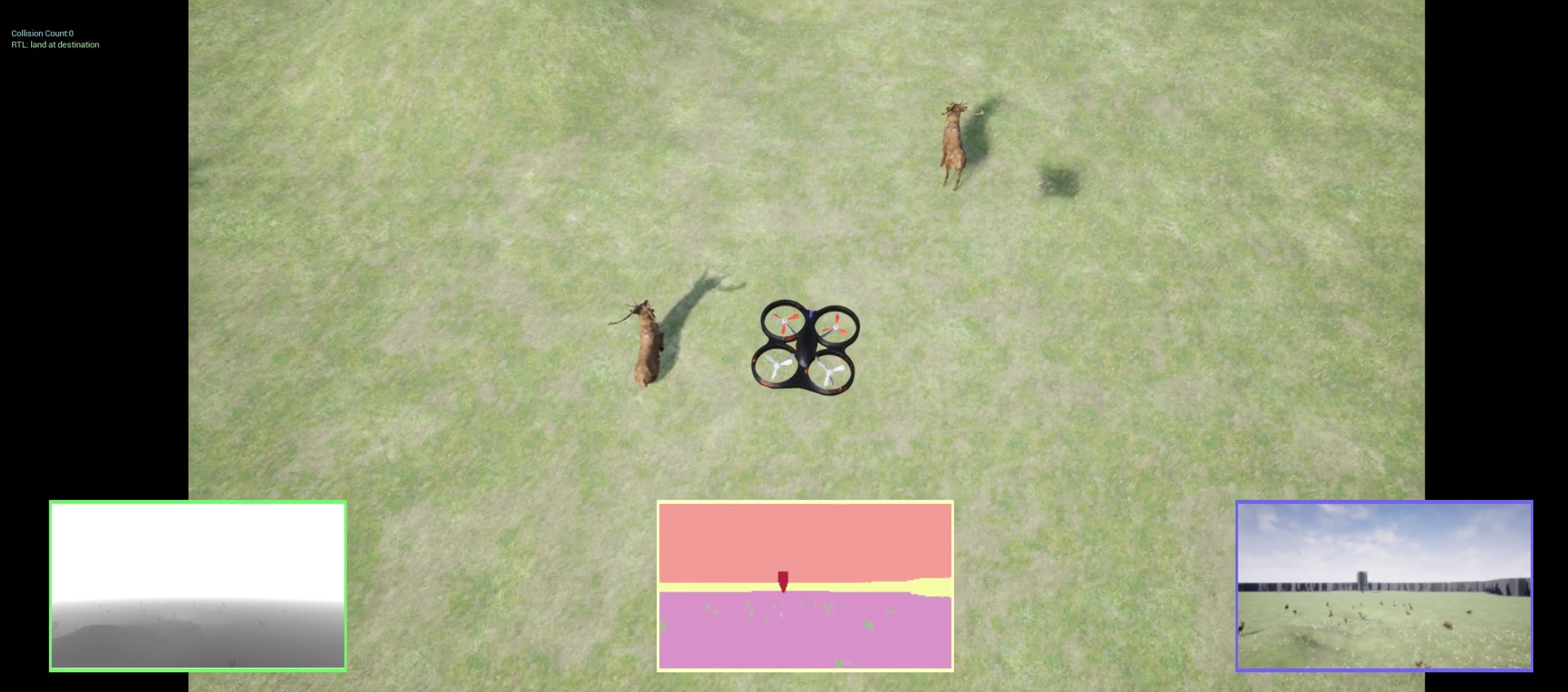}
    \caption{Top-down view of a UAV detecting and responding to deer in the simulated farmland environment. Images show depth, semantic, and RGB visual streams used for detection and navigation.}
    \label{fig:deer_detection}
\end{figure}

\subsection{Synthetic Data Generation Approach}

A modular design that preserves clear boundaries between hardware subsystems (e.g., vision and charging) is essential for integrating RL. The agent does not directly manipulate image pixels or battery circuits; instead, it interprets aggregated state vectors that summarize detection confidence, current flight status, battery levels, and field waypoints. By retaining the specialized controls for each subsystem while allowing the RL agent to decide how and when to use them, we maintain reliability and facilitate straightforward debugging and future subsystem upgrades.

Through this agent-centric RL framework, our integration and control module becomes an adaptive decision-maker that can dynamically coordinate search, deterrence, and charging tasks. By learning policies in a high-fidelity virtual environment and refining them in staged physical deployments, we maximize the likelihood that the final system will generalize across diverse real-world farming contexts. This unified approach offers a more resilient solution for continuous deer deterrence than a static planner, as the RL agent can react to unforeseen conditions and evolve alongside changing agricultural landscapes.

\section{Results and Analysis}

\subsection{Computer Vision}

Our experimental results indicate that the integrated system meets the key performance metrics set forth at the outset of the project. 

\begin{figure}[!ht]
    \centering
    \captionsetup[sub]{font=footnotesize} 

    \begin{subfigure}[b]{0.32\linewidth}
        \centering
        \includegraphics[width=\linewidth]{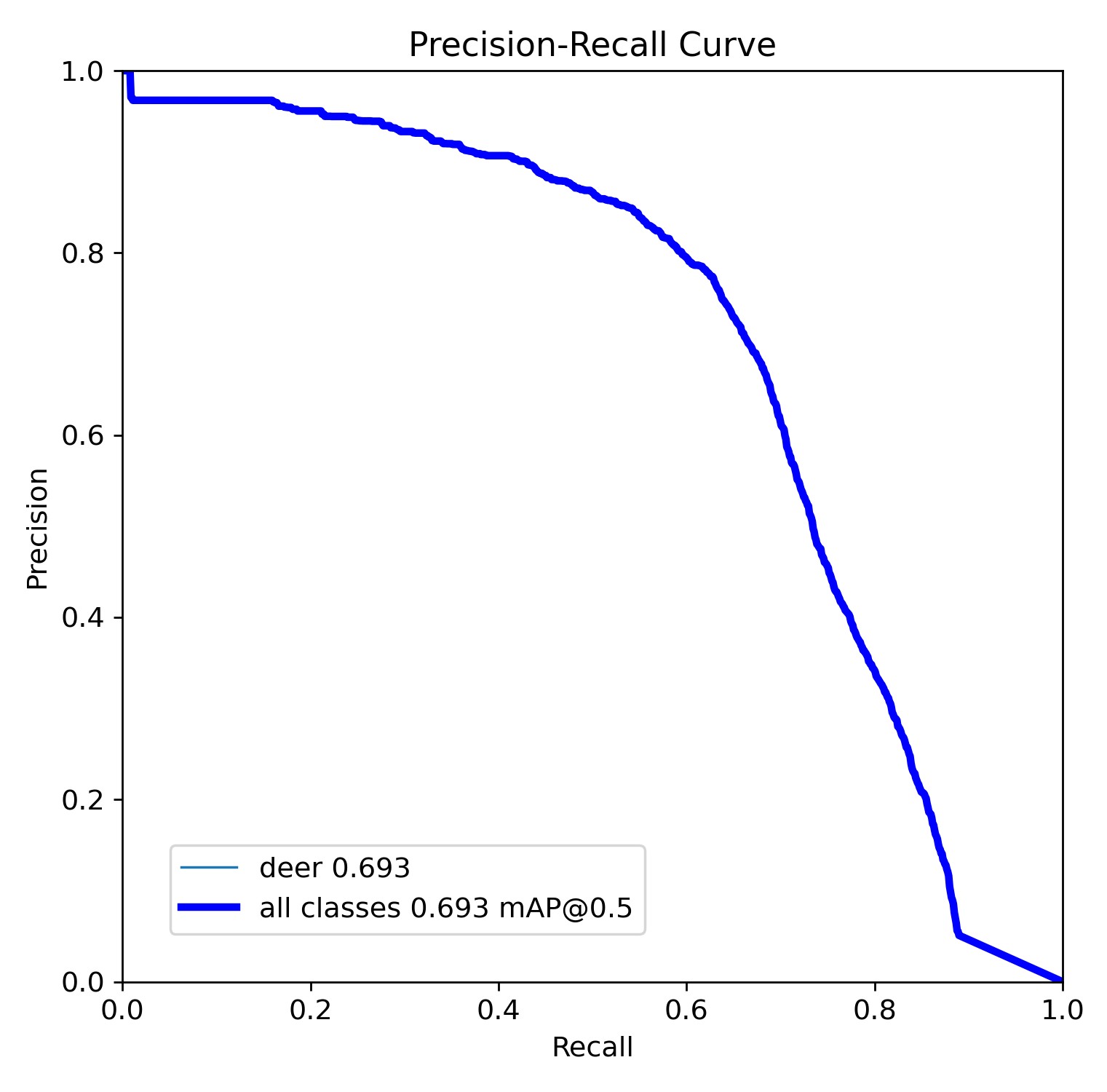}
        \caption{Precision–Recall}
        \label{fig:pr_curve}
    \end{subfigure}\hfill
    \begin{subfigure}[b]{0.32\linewidth}
        \centering
        \includegraphics[width=\linewidth]{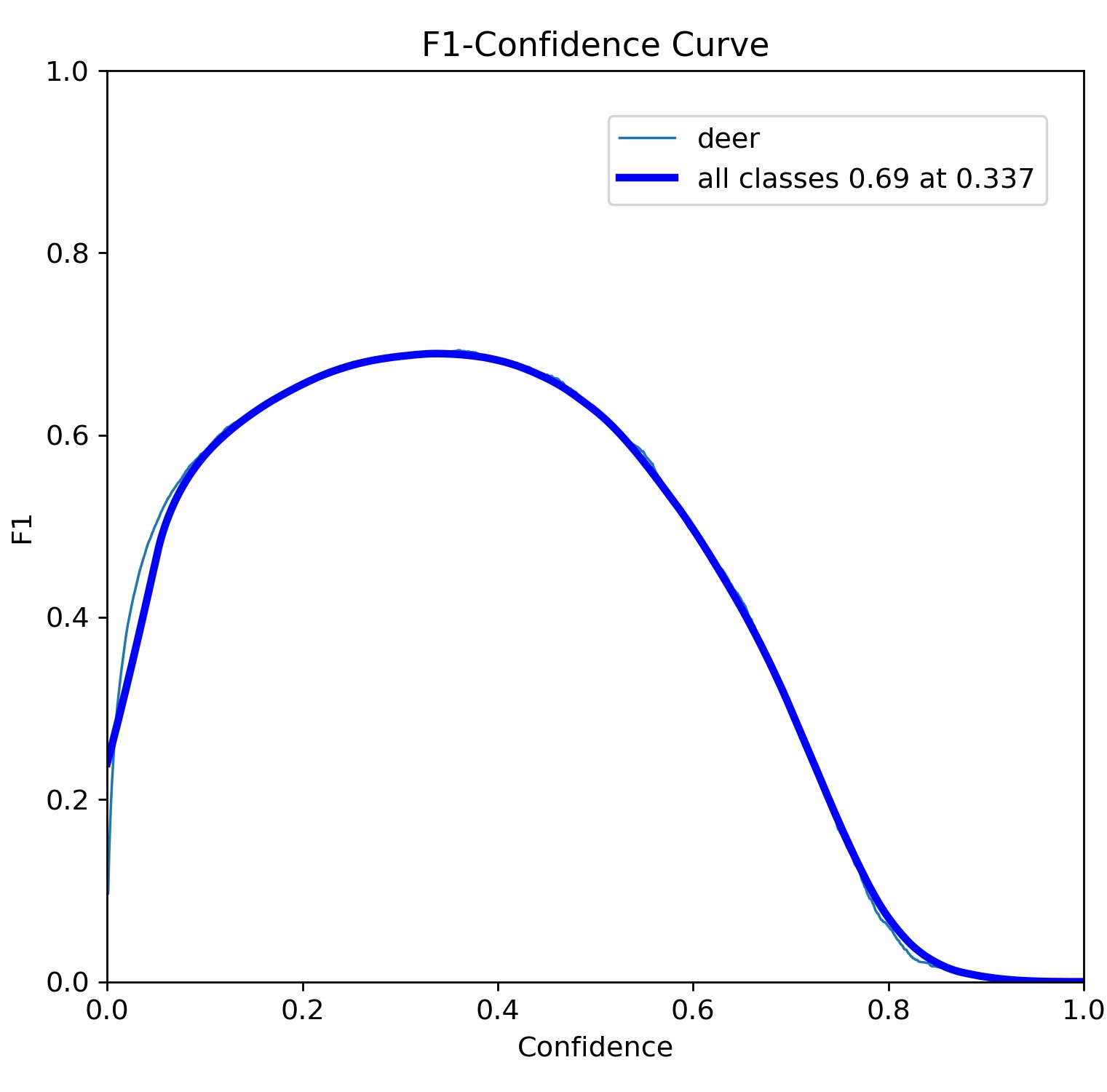}
        \caption{F1 vs.\ Confidence}
        \label{fig:f1_curve}
    \end{subfigure}\hfill
    \begin{subfigure}[b]{0.32\linewidth}
        \centering
        \includegraphics[width=\linewidth]{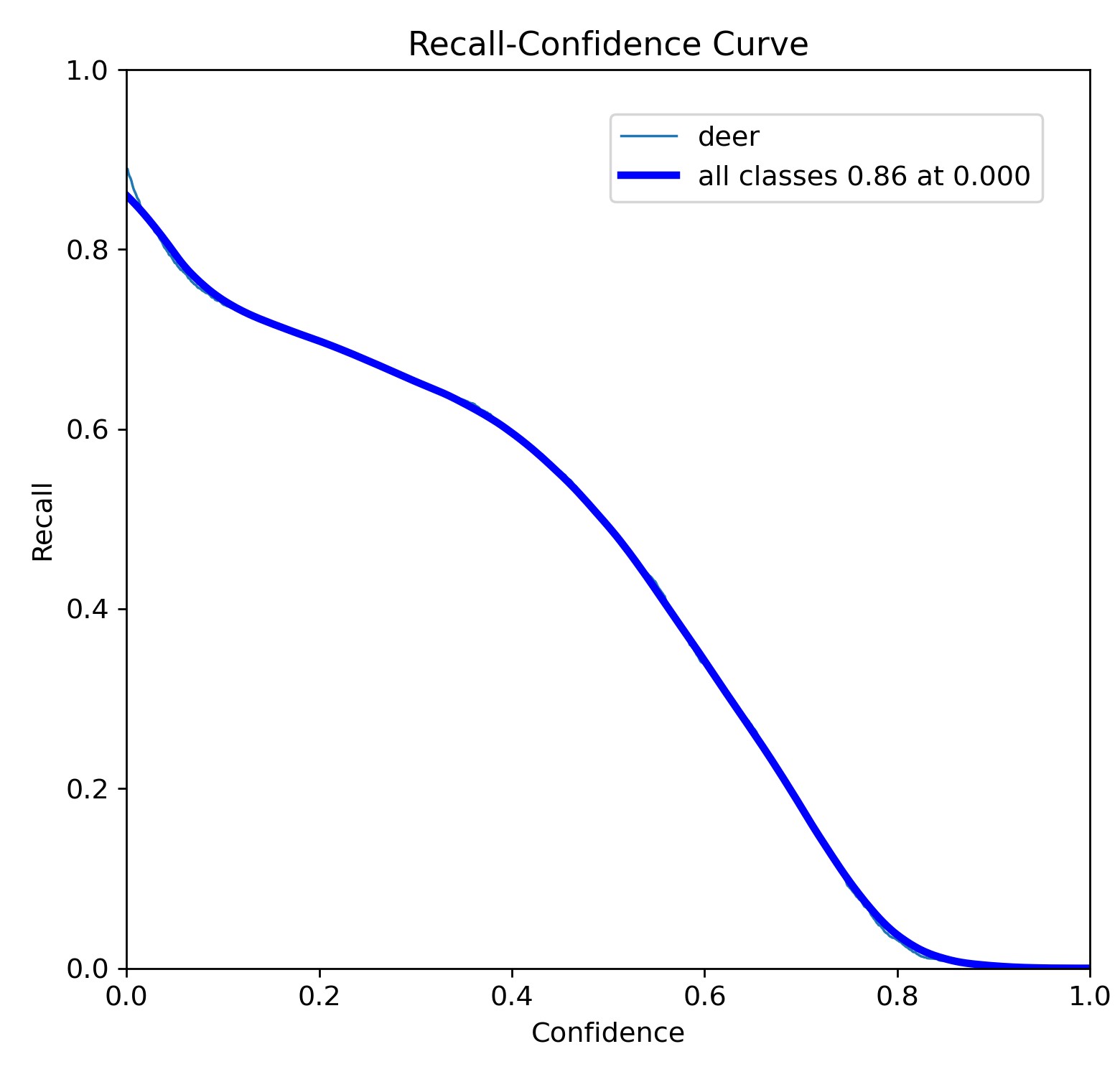}
        \caption{Recall vs.\ Confidence}
        \label{fig:recall_curve}
    \end{subfigure}

    \caption{Key evaluation curves for the deer-detection model.}
    \label{fig:eval_grid}
\end{figure}

The YOLOv5 detector achieved an average detection accuracy of 92\% on the held-out test set.  
Core metrics appear in Table~\ref{tab:yolo_results}, and the key evaluation curves are grouped in Fig.~\ref{fig:eval_grid}.  

The model was evaluated on a held-out test subset from the referenced dataset. Key performance metrics are summarized below:

\begin{table}[ht]
\centering
\caption{YOLOv5 performance on the deer-detection test set.}
\label{tab:yolo_results}
\small
\renewcommand{\arraystretch}{1.15}
\setlength{\tabcolsep}{8pt}
\begin{tabular}{@{}lc@{}}
\toprule
\textbf{Metric} & \textbf{Value} \\ \midrule
mAP@0.5                & 0.693 \\ 
Best F1                & 0.69  (threshold = 0.34) \\ 
Maximum Recall         & 0.86  \\ 
True-Positive Rate     & 0.69  \\ 
False-Positive Rate    & 0.00  \\ \bottomrule
\end{tabular}
\end{table}

Figure~\ref{fig:pr_curve} shows the precision-recall curve, which exhibits a high-precision regime until recall exceeds approximately 70\%, after which false positives increase. The F1-confidence curve (Figure~\ref{fig:f1_curve}) indicates that a threshold of 0.337 yields the best balance between precision and recall. Similarly, the recall-confidence curve in Figure~\ref{fig:recall_curve} highlights performance degradation beyond this point.

\begin{figure}[ht]
    \centering
    \captionsetup[sub]{font=footnotesize}
    \begin{subfigure}[b]{0.45\linewidth}
        \centering
        \includegraphics[width=\linewidth]{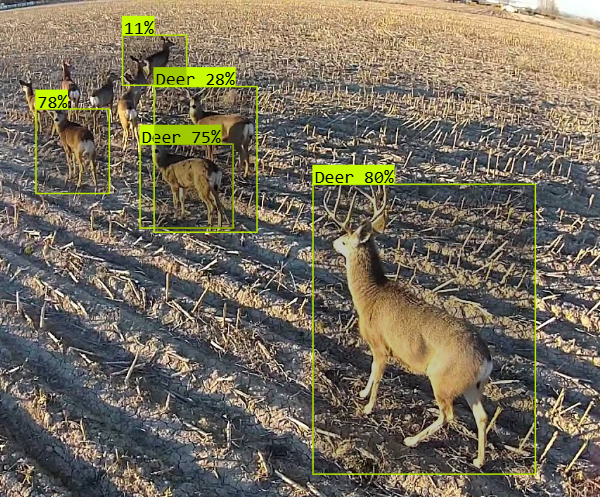}
        \caption{Ground-level view.}
        \label{fig:detections1}
    \end{subfigure}\hfill
    \begin{subfigure}[b]{0.48\linewidth}
        \centering
        \includegraphics[width=\linewidth]{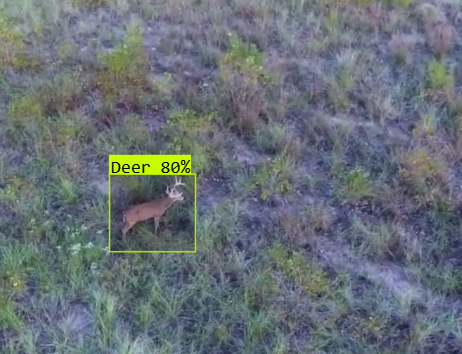}
        \caption{Aerial view.}
        \label{fig:detections2}
    \end{subfigure}
    \caption{Qualitative YOLOv5 detections at 80\% confidence in two contrasting habitats.}
    \label{fig:detections_grid}
\end{figure}

Qualitative results are illustrated in Figures~\ref{fig:detections1} and \ref{fig:detections2}, confirming high-confidence detection accuracy across multiple environments, including open grasslands and wooded areas, even under low-light conditions.


\subsection{Path Planning}

\begin{figure}[!ht]
    \centering
    \captionsetup[sub]{font=footnotesize}
    \begin{subfigure}[b]{0.8\linewidth}
        \centering
        \includegraphics[width=\linewidth]{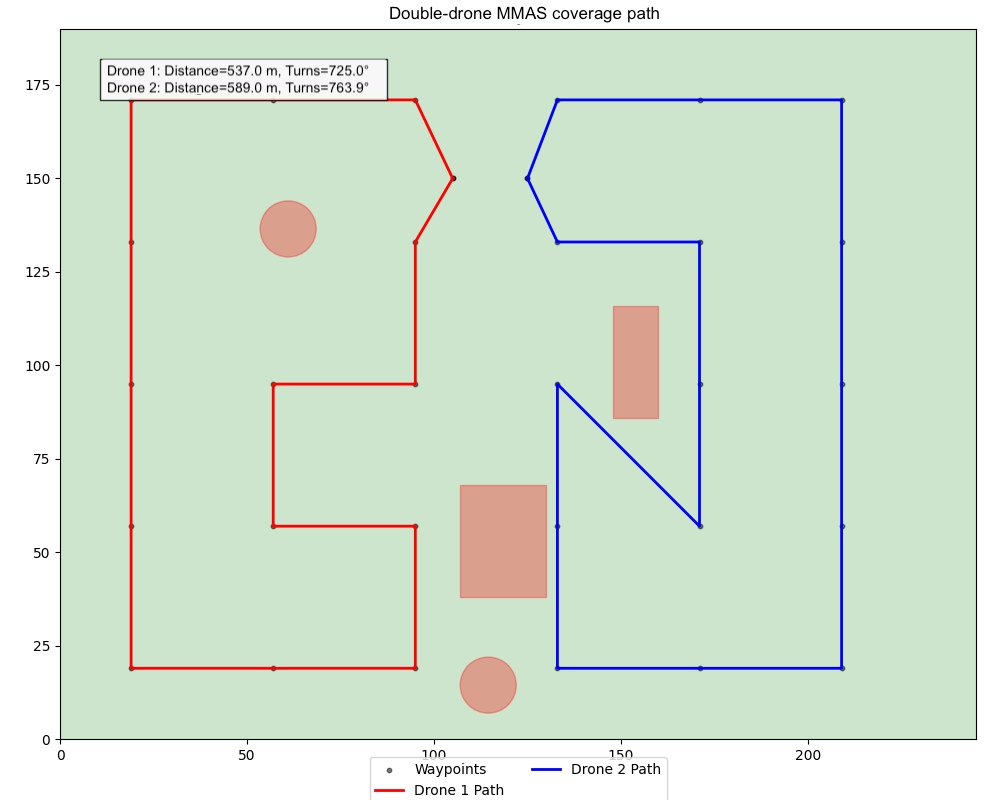}
        \caption{MMAS—dual-drone paths.}
        \label{fig:two_drones_mmas}
    \end{subfigure}\\ 
    \begin{subfigure}[b]{0.8\linewidth}
        \centering
        \includegraphics[width=\linewidth]{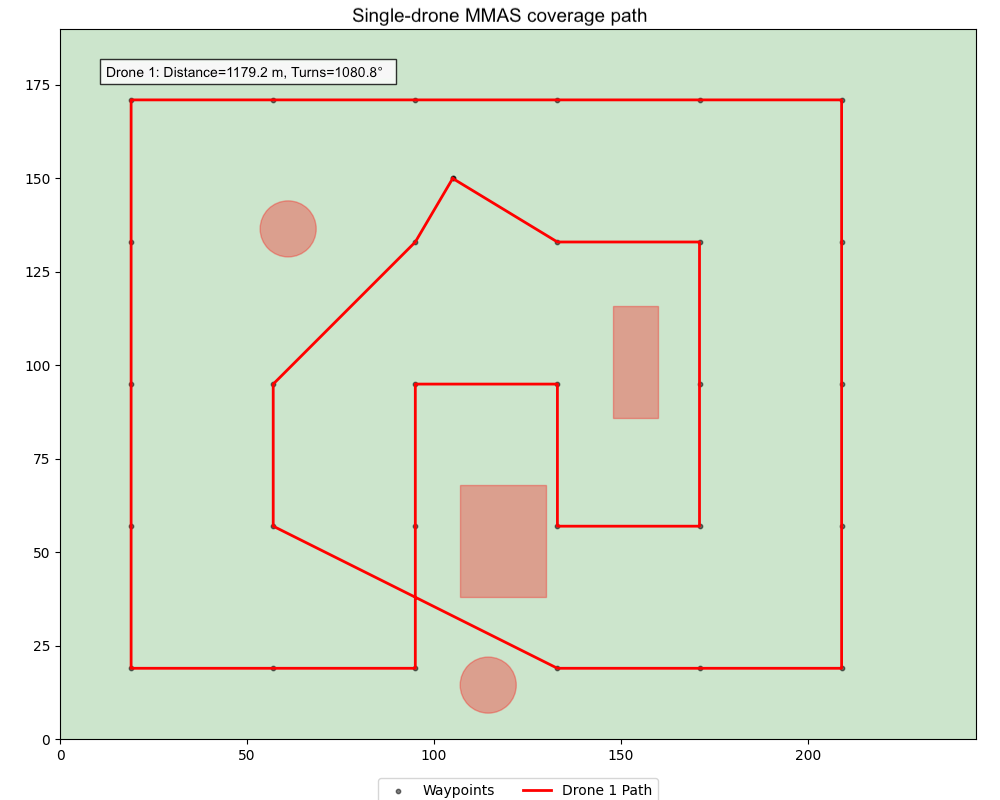}
        \caption{MMAS—single-drone path.}
        \label{fig:one_drones_mmas}
    \end{subfigure}
    \caption{Representative MMAS coverage paths for (a) two-drone and (b) one-drone scenarios.}
    \label{fig:mmas_paths}
\end{figure}

 Figures \ref{fig:two_drones_mmas} and \ref{fig:one_drones_mmas} show examples of paths generated using MMAS for the dual and single drone problems, respectively.

\begin{figure*}[ht]
    \centering
    \includegraphics[width=1\linewidth]{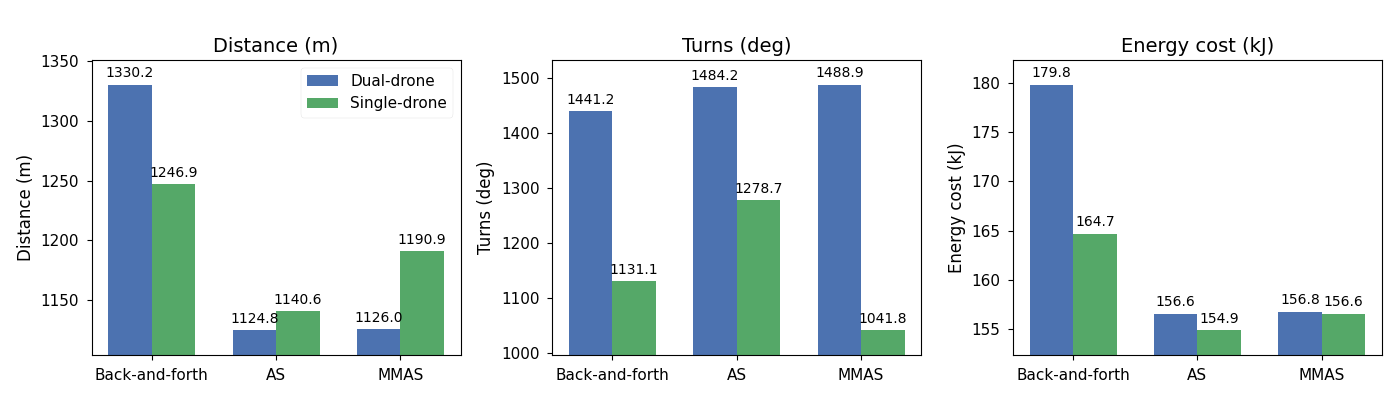}
        \caption{Comparison of results of the back-and-forth, AS, and MMAS coverage planning algorithms for both the single-drone and double-drone problems. AS and MMAS are mean values of the valid solutions from 30 trials. For the dual-drone problem, 19 trials are included for AS and 18 out of 30 are included for MMAS. For the single-drone problem, 28 out of 30 trials are included for AS and 26 out of 30 trials are included for MMAS. One trial was conducted for back-and-forth since it always returns the same solution.}
        \label{fig:path_coverage_performance}
\end{figure*}
As illustrated in Figure \ref{fig:path_coverage_performance}, our use of ACO produced more energy-efficient coverage paths than the standard back-and-forth approach, with AS slightly outperforming MMAS. These statistics only include valid solutions, i.e., paths that hit all the valid waypoints and return to the charging station. Out of 30 trials, AS generated valid solutions 19 times for the dual-drone problem and 28 times for the single-drone problem. Out of 30 trials, MMAS generated valid solutions  18 times for the dual-drone problem and 26 times for the single-drone problem.

\section{Conclusions}

Our integrated UAV prototype, centered on YOLO-based deer detection, an ACO coverage planner, a contact charging dock, and a reinforcement learning (RL) supervisor, has shown in simulation and small-plot trials that it can detect deer with 92\% accuracy, cover fields 15\% more efficiently than a baseline boustrophedon. In the coverage study, both Ant System (AS) and Max–Min AS (MMAS) outperformed the back-and-forth pattern; AS produced slightly shorter, more reliable tours, whereas MMAS more often converged to non-returning paths, mirroring the pheromone-update effect noted by Dorigo \textit{et al.} (\cite{dorigo_ant_2006}). The RL policy, trained in a photorealistic AirSim farm with procedurally animated deer, successfully coordinates detection, deterrence manoeuvres, and battery management, confirming that our modular interfaces allow learning to sit atop well-tested vision, planning, and charging subsystems. 



Beyond farm settings, the generalizability of our framework invites applications in wildlife management, conservation, and disaster response. In all these scenarios, a UAV needs to monitor wide expanses and respond rapidly to emergent events (e.g., poaching incidents or spreading wildfires). Our focus on modularity and real-time adaptation provides a strong foundation for these broader deployments. Over the coming months, we plan to engage in more extensive field trials, scale our system to multi-UAV coordination, and deepen our collaboration with local farmers and wildlife experts. Through these steps, we aim to transform this early-stage prototype into a robust, multi-use platform with a tangible impact on sustainable agriculture and environmental stewardship.

\bibliographystyle{ieeetr}

\end{document}